\documentclass[a4paper, 10pt, conference]{ieeeconf}

\IEEEoverridecommandlockouts

\usepackage{graphics} 
\usepackage{amsmath}
\usepackage{amssymb}

\usepackage{color} 
\usepackage{graphicx}
\usepackage{booktabs}
\usepackage{listings} 
\usepackage[]{nohyperref}
\usepackage{url}
\usepackage{textcomp}
\makeatletter
\let\MYcaption\@makecaption
\makeatother

\usepackage[font=footnotesize]{subcaption}

\makeatletter
\let\@makecaption\MYcaption
\makeatother

\newcommand{\lb}[1][]{\ensuremath{\llparenthesis}}
\newcommand{\rb}[1][]{\ensuremath{\rrparenthesis}}

\newtheorem{the1}{Theorem}

\title{\LARGE \bf
Acting Thoughts: Towards a Mobile Robotic Service Assistant for \\ Users with Limited Communication Skills   
}

\author{F. Burget* \quad L.D.J. Fiederer* \quad D. Kuhner* \quad M. V\"olker* \quad J. Aldinger* \quad R.T. Schirrmeister \quad C. Do \\
J. Boedecker \quad B. Nebel \quad T. Ball \quad W. Burgard \thanks{* These authors contributed equally to the work. Authors
    are with the Department of Computer Science and Faculty of Medicine, University of Freiburg, Germany.  \{burgetf, kuhnerd, do, aldinger, jboedeck, nebel, burgard\}@informatik.uni-freiburg.de and \{lukas.fiederer, martin.voelker, robin.schirrmeister, tonio.ball\}@uniklinik-freiburg.de. This research was supported by the German Research Foundation (DFG, grant number EXC
    1086) and grant BMI-Bot by the Baden-W\"urttemberg Stiftung.}}

\begin{document}

\maketitle
\thispagestyle{empty}
\pagestyle{empty}

\begin{abstract}

  As autonomous service robots become more affordable and thus available also for the general public, there is a growing need for user friendly interfaces to control the robotic system. Currently available control modalities typically expect users to be able to express their desire through either touch, speech or gesture commands. While this requirement is fulfilled for the majority of users, paralyzed users may not be able to use such systems.
  In this paper, we present a novel framework, that allows these users to  interact with a robotic service assistant in a closed-loop fashion, using only thoughts. The brain-computer interface (BCI) system is composed of several interacting components, i.e., non-invasive neuronal signal recording and decoding, high-level task planning, motion and manipulation planning as well as environment perception. In various experiments, we demonstrate its applicability and robustness in real world scenarios, considering fetch-and-carry tasks and tasks involving human-robot interaction. As our results demonstrate, our system is capable of adapting to frequent changes in the environment and reliably completing given tasks within a reasonable amount of time. Combined with high-level planning and autonomous robotic systems, interesting new perspectives open up for non-invasive BCI-based human-robot interactions.

\end{abstract}

\section{Introduction}

For patients with heavily impaired communication capabilities, such as
severly paralyzed patients, their condition forces them to constantly
rely on the help of human care-takers. Robotic service assistants can
re-establish some autonomy for these patients, if they offer adequate
interfaces and possess a sufficient level of intelligence. Generally,
an intelligent system requires adaptive task and motion planning
modules to determine appropriate task plans and motion trajectories
for the robot, that implement a task in the real world. Moreover, it
requires a perception component, e.g., to detect objects of interest
or to avoid accidental collisions with obstacles.  Typically used
interfaces, such as haptic (buttons), audio (speech) or visual
(gesture) interfaces, to command the robotic system are intuitive and
easy options for healthy users, but difficult to impossible to use for
paralyzed
individuals.

\begin{figure}[t]
	\centering
		\includegraphics[width=0.9\columnwidth]{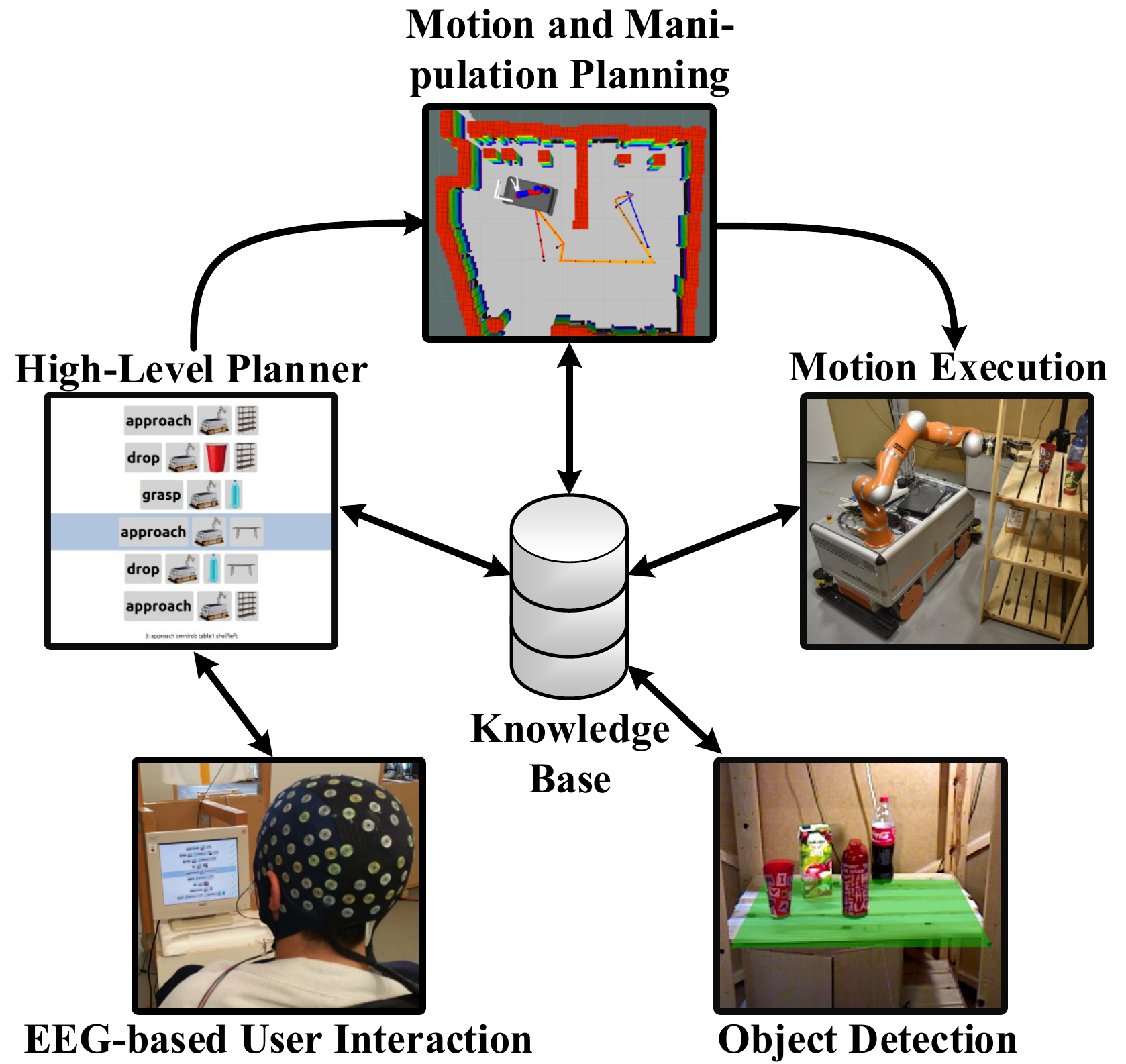}
  \caption{Framework unifying decoding of neuronal signals, high-level task planning, low-level motion and manipulation planning, scene perception with a centralized knowledge base at its core. Intuitive goal selection is provided through an adaptive graphical user interface.}
	\label{fig:introduction}
\end{figure}

In this paper, we present a novel framework, schematically depicted in
Fig.~\ref{fig:introduction}, that allows closed-loop interaction
between users with minimal communication capabilities and a robotic
service assistant. To do so, we record neuronal activity elicited in
the human brain, the common origin of all types of communication, with
an electroencephalography (EEG) system. Furthermore, we adopt a convolutional
neural network approach for online decoding of neuronal
activity, in order to allow users to navigate through a graphical user
interface (GUI) provided by a high-level task planner. The set of
feasible actions displayed in the GUI, depends in turn on the current
state of the world, which is stored in a central knowledge base and
continuously updated with information provided by the robot and a
camera perception system. Once a task has been selected, it is
decomposed into a sequence of atomic actions by the high-level
planner. Subsequently, each action is resolved to a motion for the
mobile manipulator using low-level motion and manipulation planning
techniques. In the following, the individual components shown in
Fig.~\ref{fig:introduction} will be described in detail, before
presenting a quantitative evaluation of the overall system regarding
its performance.

\section{Related Work}

Multiple previous studies have focused on robotic systems assisting
people with disabilities. For example, Park \emph{et
  al.}~\cite{park2016towards} implemented a system for the autonomous
feeding of yogurt to a person.  Chung \emph{et
  al.}~\cite{chung2013autonomous} focused on autonomous drinking which
involved locating the drink, picking it up and bringing it to the
person's mouth.  Using a hybrid BCI and head movement control, Achic
\emph{et al.}~\cite{achic2016hybrid} studied a setup with a moving
wheelchair and an attached robotic arm. None of these systems used
pure BCI control.  In contrast, Wang \emph{et
  al.}~\cite{wang2011motor} used a motor imagery BCI with three
classes to achieve low-level control of a robotic arm.  More relevant,
Schr\"oer \emph{et al.}~\cite{schroer2015autonomous} developed a
robotic system which receives a BCI command from a user and
autonomously assists the user in drinking from a cup. However, this
approach only considers a single object and a fixed-base manipulator.
More recently, Muelling \emph{et al.}~\cite{Muelling2017} presented a
shared-control approach to assistive robotics, albeit focused on
invasive BCIs. Nonetheless, their approach could be combined with the
high-level planning approach presented in our work.

In these applications, robust decoding of brain signals is required. Inspired by the successes of deep convolutional neural networks (ConvNets) in computer vision~\cite{krizhevsky_imagenet_2012, he_deep_2015} and speech recognition~\cite{sainath_deep_2015, sercu_very_2016}, deep ConvNets have recently been applied more frequently to EEG brain-signal decoding. Deep ConvNets were already applied to decoding tasks useful for building brain-computer interfaces.  Lawhern \emph{et al.}~\cite{DBLP:journals/corr/LawhernSWGHL16} used a deep ConvNet to decode P300 oddball signals, feedback error-related negativity and two movement-related tasks.  When evaluated cross-subject, i.e., trained on some subjects and evaluated on others, their ConvNet yields competitive accuracies compared with widely-used traditional brain-signal decoding algorithms.  Tabar and Halici~\cite{tabar_novel_2017} used a ConvNet combined with a convolutional stacked auto-encoder to decode motor imagery within-subject, yielding better accuracies than several non-ConvNet decoding algorithms.  Schirrmeister \emph{et al.}~\cite{schirrmeister2017deep} used a shallow and a deep ConvNet to decode both motor imagery and motor execution within-subject, reaching or slightly surpassing the accuracies of the widely used EEG motor-decoding algorithm \textit{filter bank common spatial patterns}~\cite{ang_filter_2008}.  Bashivan \emph{et al.}~\cite{bashivan_learning_2016} used a ConvNet trained on fourier-transformed inputs to estimate mental workload.  In addition to the work on evaluating ConvNet decoding accuracies, ConvNet visualization methods allow us to get a sense of what brain-signal features the network is using~\cite{schirrmeister2017deep, bashivan_learning_2016, stober_learning_2016}. Taken together, these advances make deep ConvNets a viable alternative for brain-signal decoding in brain-computer interfaces.  Still, to our knowledge, online control with deep ConvNets has not yet been reported for an EEG-based brain-computer interface.

\section{Online Decoding of Neuronal Signals}
\label{sec:neuronal_control}

The system at hand is developed to control more complex scenarios than the ones considered in previous work. Particularly, we consider scenarios
involving manipulation of objects as well as human-robot
interaction. Feasible goals are determined by our GUI which is
controlled by directional commands. As reliable classification of
brain signals into navigation directions cannot yet be achieved directly
with non-invasive BCIs, we used a deep ConvNet approach for decoding of
multiple mental tasks from EEG (Schirrmeister \emph{et
  al.}~\cite{schirrmeister2017deep}). This approach introduces a hybrid
network, combining a deep ConvNet with a shallower ConvNet
architecture. The deep part consists of 4 convolution-pooling blocks
using exponential linear units (ELU)~\cite{clevert_fast_2016} and max
pooling, whereas the shallow part uses a single convolution-pooling
block with squaring non-linearities and mean pooling. Both parts use a
final convolution with ELU to produce output features. These features
are then concatenated and fed to a final classification layer. We
trained the ConvNet to decode five mental tasks: right hand finger and both feet toe movements, object rotation, word generation
and rest. These mental tasks evoke discernible brain patterns and are
used as surrogate signals to control the GUI. \emph{Offline}
training was done with a cropped training strategy using shifted time
windows within the trials as input data~\cite{schirrmeister2017deep}.

From our experience it is important to train the BCI decoder and
subjects in an environment that is as close as possible to the real
application environment to avoid pronounced performance drops. 
Therefore, we designed a gradual training paradigm within the high-level planner GUI
where the displayed environment, timing and actions are identical to
those of the real control task. The training paradigm proceeds as
follows: We first train each subject \emph{offline} using simulated
feedback. Subjects are aware of not being in control of the GUI. The
mental tasks are cued using grayscale images presented for 0.5\,s in
the center of the display. At all times a fixation circle is displayed
at the center of the GUI and the subject is instructed to fixate on it
to minimize eye movements. After a random time interval of 1-7\,s the
fixation circle is switched to a disk for 0.2\,s, which indicates the
end of the mental task. At the same time the GUI action (\textit{go
  up, go down, select, go back, do nothing}) corresponding to the cued
mental task is performed to update the GUI. 
To keep training realistic we include a 20\,\% error rate, i.e., on
average every fifth action is erroneous. We instruct the subjects to
count the error occurrences to assert their vigilancy. This
offline data is used to train the individual deep ConvNets. Then, the subjects do
\emph{online} training by performing the decoded mental tasks in the
GUI. Finally, we stop cueing the mental tasks. To evaluate the
performance of the BCI control, we let the subjects create instructed
high-level plans in the GUI. These tasks are then executed by a
simulated robot or the real mobile manipulator, when available. To
provide more control over the mobile manipulator and enhance the
feeling of agency, subjects have to confirm the execution of every
planned action and can interrupt the chain of actions at any moment
during their execution. BCI decoding accuracies for the label-less
instructed tasks are assessed by manually rating each decoding based
on the instructed task steps. Statistical significance of the decoding
accuracies were tested using a conventional permutation test with 100\,k random permutations of the labels (i.e., 
p-value is the fraction of label permutations that would have led to better or equal accuracies than the accuracy for the original labels).

\section{High-Level Goal Formulation Planning}

We use domain independent planning to derive the required steps for
reaching a desired high-level goal in a complex task. The user can
formulate a high-level goal without knowledge of the internal
representation of objects in the planning system and the exact
capabilities of the robot. This is achieved by an intuitive graphical
user interface, where the object parameters of the goal are
specified by incrementally refining the objects by referring to their
\emph{type}, e.g., ``cup'' or \emph{attributes}, e.g., ``content =
apple-juice''.

Domain independent planning identifies a sequence of actions that
transforms the current world state into a state satisfying a goal
condition. A planning task consists of: (i) a planning domain
describing static components such as the object type hierarchy and the
available actions and (ii) a problem instance describing the objects
present in the world and their current state, as well as a goal
description. While the current state of the objects can be extracted
from the knowledge base, the goal has to be chosen in the GUI.

A restricted vocabulary is shared between the user and the planning
system. Objects or sets of objects are identified by creating
\emph{referring expressions} to them composed of \emph{shared
  references} built on this vocabluary~\cite{dale1995}. We briefly
describe the relevant aspects of our previous work in this
area~\cite{goebelbecker}.  In general, a \emph{referring expression}
$\phi$ is a logical formula with a single free variable. $\phi$
\emph{refers} to an object $o$ if $\phi(o)$ is valid. E.g., the
reference $\phi(x) \equiv \mathit{cup}(x) \land \mathit{contains}(x,
water)$ refers to all cups containing water. We restrict ourselves to
references that are simple conjunctions of facts, which is not only
preferable for computational reasons, but also allows us to
incrementally refine references by adding constraints. For example,
adding $\mathit{contains}(x, water)$ to $\mathit{cup}(x)$, restricts
the set of all cups to the set of cups containing water.

We distinguish between three types of fundamental object references:
\emph{individual references}, \emph{typename references} and
\emph{relational references}. \emph{Individual references} are
identified by name, such as the ``omniRob'' robot. \emph{Typename
  references} can be identified by the name of their type. While we
cannot refer to the cups in our scenario directly, we can refer to an
unspecific cup.  \emph{Relational references} are encountered when objects can be
referred to via predicates in which they occur as an argument. The
relations in our scenario are object attributes. For example, the
\textit{content} of the cup is used to clarify which cup is
meant. These object references are used to create references to
goals. We start defining goals with the action that achieves it as we
found that this is most natural to the user, e.g., $\mathit{put}(x,y)
\land \mathit{cup}(x) \land \mathit{shelf}(y)$.  After the initial
selection of a goal type (e.g., \textit{drop}) it is necessary to
determine the objects for all parameters of the goal predicate or
action. These parameters are refined by constraining the previous
choice until the argument is either determined uniquely (i.e., it is
impossible to constrain the argument further) or the user declares
that any remaining option is acceptable. We exclude unreachable goals,
but we allow for goals that can only be achieved after a sequence of
preceding actions (e.g., \textit{drinking water} could require to
fetch a cup, bring it to the patient, fetch a bottle and pour the
water into the cup in order to be executed). The goal that is
determined by the selection process of the GUI is then passed to a
custom domain independent planner.

\section{Robot Motion Generation}
\label{sec:robot_motion_generation}

For generating paths for the mobile base, we apply the sampling-based
planning framework \textit{BI$^2$RRT*}~\cite{burget16iros}. Given a
pair of terminal configurations, it performs a bidirectional search
using uniform sampling in the configuration space until an initial
sub-optimal solution path is found. This path is subsequently refined
for the remaining planning time, adopting an informed sampling
strategy, which yields a higher rate of convergence towards the
optimal solution. Execution of paths is implemented via a closed-loop
joint trajectory tracking algorithm using robot localization feedback.

To realize pick, place, pour and drink motions efficiently, we adopt a 
probabilistic roadmap planner approach~\cite{kavraki1998probabilistic}. 
The planner uses a graph of randomly sampled task poses (end-effector poses), which are
connected by edges. To find a plan between two poses, the planner
connects both poses with the roadmap graph and uses the A* algorithm to find an
optimal path between the start and goal pose. The execution of the
plan maps the task space velocity commands to joint velocity commands
by employing a task space motion controller. We sample random poses around the object to determine grasp motions. For dropping objects we extract horizontal planes from the camera's point cloud and sample poses above those planes to find a suitable drop location. 

\section{Implementation Details}

Implementation of our framework in the real world requires several components, such as neuronal signal decoding, scene perception, knowledge base operations as well as symbolic and motion planning, to run in parallel. Therefore, we distributed the computation across a network of 7 computers, communicating among each other via ROS. The decoding of neuronal signals has four components. EEG measurements are performed using \textit{Waveguard EEG} caps with 64 electrodes and a \textit{NeurOne} amplifier in AC mode. Additionally, vertical and horizontal EOGs, EMGs of the four extremities and ECG's are recorded. For recording and online-preprocessing, we used BCI2000 
and Matlab. We then transferred the data to a GPU server where our deep ConvNet classified the data into 5 classes. 
The high-level planner GUI consists of a back- and front-end. The
back-end of the GUI uses the \textit{Fast~Downward}
planner~\cite{helmert} to iteratively build goal references and to
find symbolic plans for the selected goal. As the planning time is not
crucial for the performance of our system, we used
\textit{Fast~Downward} with a basic configuration in our experiments.
The central knowledge base is implemented as a ROS node, which is able
to store objects with arbitrary attribute information. All changes in the
knowledge base automatically trigger updates of the front-end, unexpected ones 
interrupt the current motion trajectory execution.
Finally, we used \textit{SimTrack}~\cite{pauwels2015simtrack} for object pose detection and tracking. 

\begin{figure}[t]
  \centering
  \includegraphics[width=0.75\columnwidth]{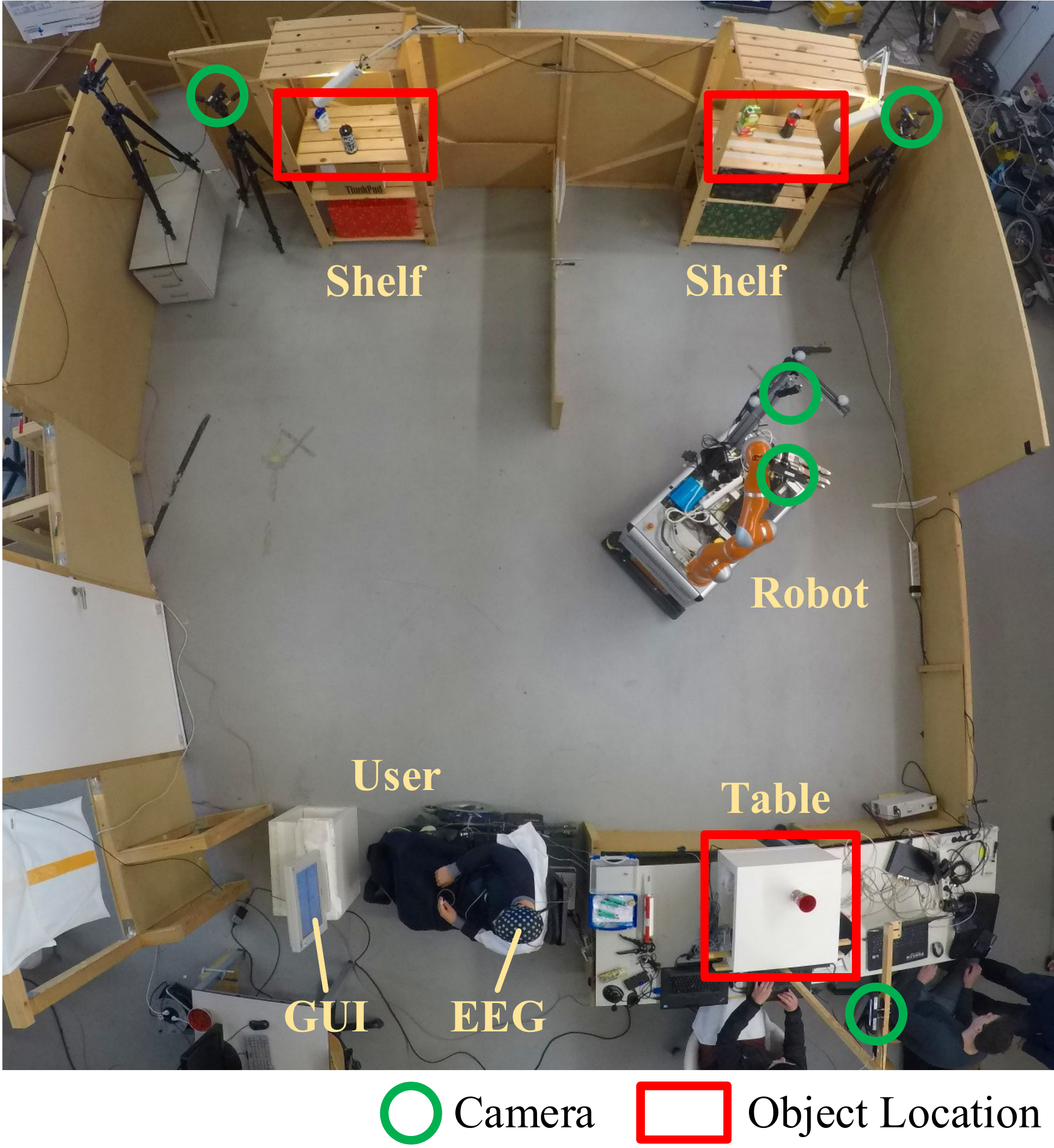}
  \caption{Experimental environment: Two shelves and a table can
    be considered by the robot for performing manipulation actions. Five RGBD sensors observe the
    environment. A human operator selects a goal using EEG control and
    the high-level planner GUI.}
  \label{fig:experimental_setup}
\end{figure}

\section{Experiments}

To evaluate our framework, we consider the environment schematically depicted in Fig.~\ref{fig:experimental_setup}, containing two shelves and a table as potential locations for manipulation actions. The user sits in a wheelchair in front of a screen, displaying the graphical interface of the high-level planner. The robot used in the experiments is the \textit{omniRob} omni-directional mobile manipulator platform by KUKA Robotics, which is composed of 10 degrees of freedom (DOF), i.e., 3 DOF for the mobile base and 7 DOF for the manipulator. Additionally, the \textit{Dexterous Hand 2.0} by Schunk is attached to the manipulator's flange and used to perform grasping and manipulation actions. The tasks we considered in our experiments required the robotic system to autonomously perform the following actions: \textit{drive from one location to another}, \textit{pick up an object}, \textit{drop an object (on a shelf or table)}, \textit{pour liquid from a bottle into a cup}, \textit{supply a user with a drink}.
Moreover, we use a perception system composed of five RGBD cameras. Three of them are statically mounted at the shelves and the table, in order to observe the scene and to report captured information to the knowledge base. The other two cameras are carried by the robot on-board. The first one is located at the mobile base and used to perform collision checks in manipulation planning. The second camera is mounted at the robot's end-effector and used for tasks involving physical human-robot interaction. A demonstration of our framework can be found in the accompanying video: \href{http://www.informatik.uni-freiburg.de/~burgetf/ecmr17/}{http://www.informatik.uni-freiburg.de/\texttildelow burgetf/ecmr17/}.

\subsection{Online Decoding of Neuronal Signals}\label{sec:bci}
We evaluated the BCI control setup with four healthy subjects (S1-4, all right-handed, three  females, aged 26.75$\pm$5.9). At the time of writing the validation, S4 was still in progress and no validation with the mobile manipulator was performed. In total, 52 runs have been recorded (20 with the real robot) where the subjects executed various instructed high-level plans. For 32 runs, we used simulated feedback from the GUI in order to generate a significant amount of data for the evaluation. 
The performance of the BCI decoding during these runs was assessed using video recordings of interactions with the GUI. We rated GUI actions as correct if they correspond to the instructed path and incorrect otherwise. Actions which are necessary to remediate a previous error are interpreted as correct if the correction is intentionally clear. Finally, we rated \textit{rest} actions as correct during the (simulated) robot executions, incorrect if the next robotic action had to be initialized and ignored them during high-level plan creation. For evaluation, five metrics have been extracted from the video recordings: (i) the accuracy of the control, (ii) the time it took the subjects to execute a high-level plan, (iii) the number of steps used to execute a high-level plan, (iv) the path optimality, i.e., the ratio of the steps used to the minimally possible number of steps, and (v) the average time per step. We summarized the results in Table~\ref{tab:exp_bci}. In total, a 76.67\,\% correct BCI control was achieved, which required 9\,s per step. Selecting a plan using the GUI took on average 148\,s and required the user to perform on average 16.74 steps in the GUI of the high-level planner. The path formed by these steps is on average 34.6\,\% away from the optimal path. The decoding accuracy of every subject is significantly above chance ($\text{p}<10^{-6}$).

\begin{table}
	\begin{center}
          \caption{Aggregated mean$\pm$std results for 52 BCI control runs (Exp.~\ref{sec:bci}), * p-value $< 10^{-6}$ \label{tab:exp_bci}}
          \setlength\tabcolsep{2pt}
		\begin{tabular}{lccccccccccc}
		    \toprule
		    & \textbf{Runs} & \textbf{Accuracy*} & \textbf{Time} & \textbf{Steps} & \textbf{Path Optimality} & \textbf{Time/Step}\\
		    & \textbf{\#}	& \textbf{[\%]} & \textbf{[s]} & \textbf{\#} & \textbf{[\%]} & \textbf{[s]}\\
		    \midrule
		    S1 	& 18	&	 84.1$\pm$6.1 	& 125$\pm$84		& 13.0$\pm$7.8 		& 70.1$\pm$22.3 	& 9$\pm$2\\
		    S2 	& 14	&	 76.8$\pm$14.1 	& 150$\pm$32		& 10.1$\pm$2.8		& 91.3$\pm$12.0	& 9$\pm$3\\
		    S3	& 17	&	 82.0$\pm$7.4	& 200$\pm$159		& 17.6$\pm$11.4		& 65.7$\pm$28.9 	& 11$\pm$4\\
		    S4	& 3		&	 63.8$\pm$15.6	& 176$\pm$102		& 26.3$\pm$11.2		& 34.5$\pm$1.2 	& 6$\pm$2\\
		    \midrule
		    	& 52	& 	76.7$\pm$9.1	& 148$\pm$50 	& 16.7$\pm$7.1		& 65.4$\pm$23.4 	& 9$\pm$2\\
		    \bottomrule
  		\end{tabular}
	\end{center}
\end{table}

\begin{figure}
	\begin{center}
		\includegraphics[width=1\columnwidth]{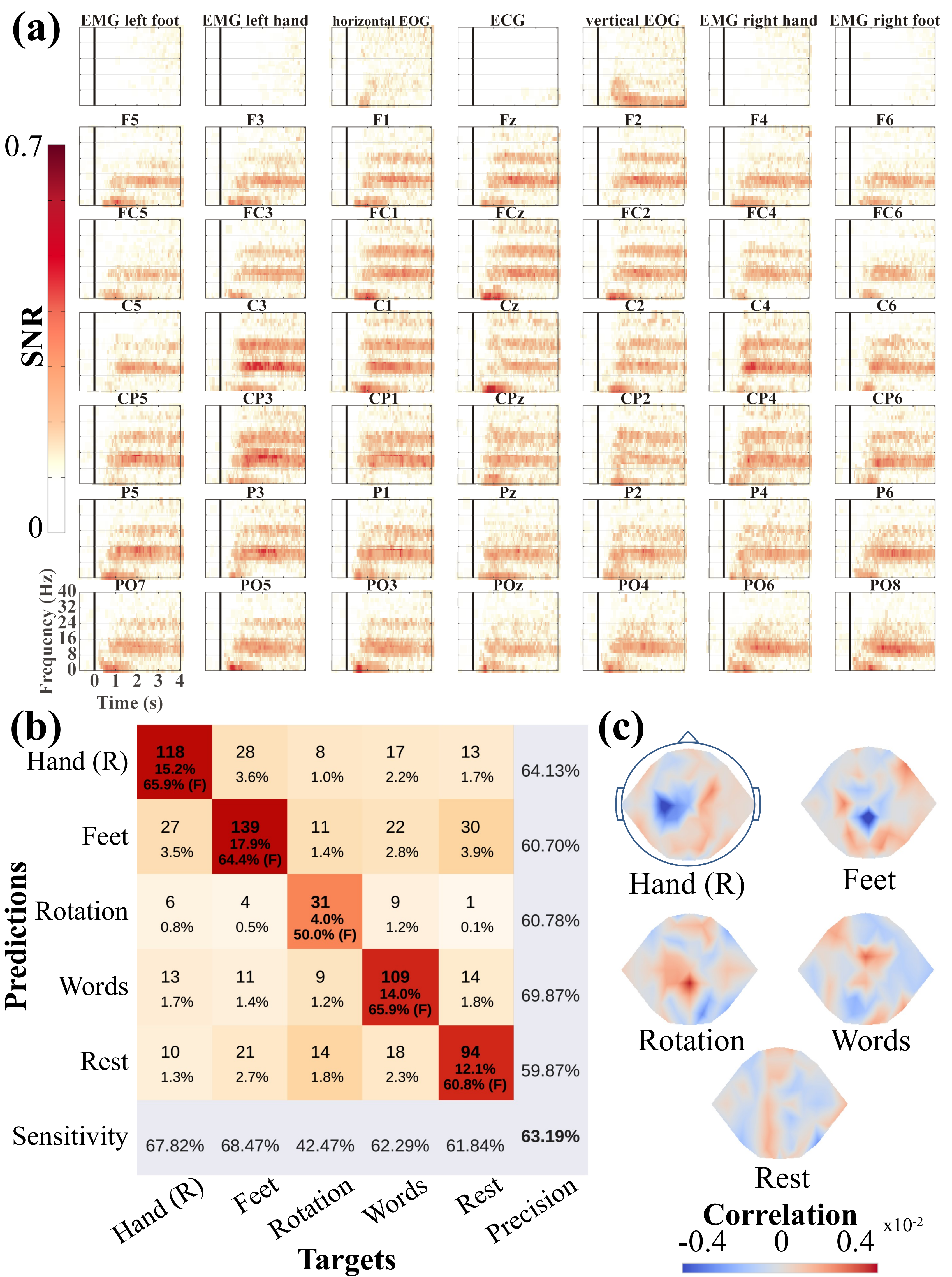}
		\caption{EEG data and decoding results. \textbf{(a)} SNR of the first 4\,s of data used to train the hybrid ConvNet. Highest SNR can be observed in the alpha (7-14\,Hz) and lower beta (16-26\,Hz) bands. These frequency bands are robust markers of task related mental tasks. Note that the non-EEG channels (top row) were withheld from the ConvNets at any time and are displayed as negative control. Not all channels are displayed because of space constraints. \textbf{(b)} Confusion matrix of decoding accuracies for the train/test transfer. Accuracies are well above the theoretical chance level of 20\,\%. \textbf{(c)} Topographically plausible input-perturbation network-prediction correlation maps in the alpha (7-13 Hz) frequency band. For details on the visualization technique we refer the reader to~\cite{schirrmeister2017deep}. }
	\label{fig:training_data}
	\end{center}
\end{figure}

The subject-averaged EEG data used to train the hybrid ConvNets and the decoding results of the train/test transfer are visualized in Fig.~\ref{fig:training_data}. In Fig.~\ref{fig:training_data}(a) we show the signal-to-noise ratio (SNR) of all 5 classes $\mathcal{C}$ of the labeled datasets. We define the SNR for a given frequency $f$, time $t$ and electrode $e$ as
\begin{equation*}
\text{SNR}_{f,t,e} = \frac{\text{IQR}\left(\left\{\text{median}\left(\mathcal{M}_i\right)\right\}\right)}{\text{median}\left(\left\{\text{IQR}\left(\mathcal{M}_i\right)\right\}\right)} \quad i \in \mathcal{C},
\end{equation*}
where $\mathcal{M}_i$ corresponds to the set of values at position $(f, t, e)$ of the $i$-th task, with $|\mathcal{M}_i|$ being the number of repetitions. $\text{median}(\cdot)$ and $\text{IQR}(\cdot)$ is the median and interquartile range (IQR), respectively. The upper part describes the variance of the class medians, i.e., a large variance means more distinguishable class clusters and a higher SNR. The denominator describes the variance of values in each class, i.e., a lower variance of values results in a higher SNR. The low SNR in EMG channels shows that the subjects did not move during the tasks.

The decoding accuracies achieved on the test data after initial training of the ConvNets are visualized in Fig.~\ref{fig:training_data}(b). To further support the neural origin of the BCI control signals, Fig.~\ref{fig:training_data}(c) shows physiologically plausible  input-perturbation network-prediction correlation results (see~\cite{schirrmeister2017deep} for methods).

\subsection{Fetch and Carry Task} \label{sec:exp_fetch_carry}
The first experiment, considering the use of the real robot, evaluates the complete system in fetch-and-carry tasks. The goal was to transfer an object from one location to another, e.g.,  from a shelf to the table, using the robot. To fulfill such tasks the robot typically needs to execute four subtasks: \textit{approach} object location, \textit{grasp} object, \textit{approach} other location, \textit{drop} object. The user was instructed to select a pre-defined goal using the EEG-controlled high-level planner. Moreover, we selected a random initial placement for the objects in each run, in order to cover different environment states. The experiment was repeated ten times by the user. Table~\ref{tab:exp_fetch} shows the averaged results for the experiment. The second column indicates the overall number of desired action calls, as scheduled by the high-level planner, as well as the number of calls actually performed. The third to fifth columns represent the success rate, mean and standard deviation for the runtime of actions, respectively.
Note, that the number of scheduled and actually executed actions might differ for two reasons. A number of executed calls, lower than the scheduled ones, indicates that a previous action step has failed to succeed and plan recovery was not possible. On the other hand, a higher number of executed calls indicates that the user was able to achieve plan recovery by commanding a repetition of the failed action. Moreover, we recorded the largest standard deviation for the \textit{approach} action, which can be attributed to the diverse complexity of the planning problem for the mobile base and the distance to travel between the selected grasp and drop location. In total, our system achieved a success rate of 80\% for the entire task. Planning and execution required on average 140.63$\pm$36.7\,s. Errors were mainly caused by object detection issues, i.e., the system was not able to detect the object or the detection was not precise enough to be able to successfully grasp or drop an object. 

\begin{table}
	\begin{center}
          \caption{Aggregated results for 10 runs (Exp.~\ref{sec:exp_fetch_carry})\label{tab:exp_fetch}}
		\begin{tabular}{lcccc}
		    \toprule
		    \textbf{Actions} & \textbf{\# Executions} & \textbf{Success} & \multicolumn{2}{c}{\textbf{Runtime [s]}}\\
		    					& \textbf{(\# Scheduled)} & \textbf{Executions [\%]} & \textbf{Mean}		& \textbf{Std}							\\
		    \midrule
		    Grasp 			& 10 (10) &  90.0		& 37.56		& 4.62				\\
		    Drop 			& 9 (10)  &  89.0		& 34.13		& 5.75				\\
		    Approach			& 19 (20) & 100.00		& 33.05		& 18.48				\\
		    \midrule
		    Total			& 38 (40) & 94.74		& 34.42	& 14.02					\\
		    \bottomrule
  		\end{tabular}
	\end{center}
\end{table}

\begin{table}
	\begin{center}
          \caption{Aggregated results for 10 runs (Exp.~\ref{sec:exp_drinking})\label{tab:exp_drink}}
		\begin{tabular}{lcccc}
		    \toprule
		    \textbf{Actions} & \textbf{\# Executions} & \textbf{Success} & \multicolumn{2}{c}{\textbf{Runtime [s]}}\\
		    					& \textbf{(\# Scheduled)} & \textbf{Executions [\%]} & \textbf{Mean}		& \textbf{Std}							\\
		    \midrule
		    Grasp 			& 34 (30) 	&  91.0			& 40.42		& 10.31				\\
		    Drop 			& 30 (30) 	&  97.0			& 37.59		& 4.83				\\
		    Approach			& 80 (80) 	&  100.0		& 20.91		& 7.68
		    				    \\
		    Pour				& 10 (10) 	&  100.0		& 62.90		& 7.19
		    					\\
		    Drink			& 13 (10) 	&  77.0			& 57.10		& 8.20
		    					\\
		    \midrule
		    Total			& 167 (160) & 95.86		& 32.46		& 15.51				\\
		    \bottomrule
  		\end{tabular}
	\end{center}
\end{table}

\subsection{Drinking Task} \label{sec:exp_drinking}
The last experiment evaluates the direct interaction between user and robot. Therefore, we implemented an autonomous robotic drinking assistant. Our approach enables the robot to fill a cup with a liquid, move the robot to the user and finally provide the drink to the user by execution of the corresponding drinking motion in front of the user's mouth. In addition to the techniques described above, successful pouring and drinking using a robot requires the detection of the liquid level in the cup and a reliable detection and localization of the user's mouth.

To detect the liquid level while pouring, we follow a vision-based approach introduced by Do \emph{et al.}~\cite{do16iros}. Given the camera's viewing angle and the liquid's index of refraction, the liquid height is determined from the depth measurement using a relationship based on Snell's law (see~\cite{hara2014iros} for more details). Using this knowledge, we first detect the cup, extract the depth values for the liquid and finally estimate the real liquid height. The type of liquid and hence the index of refraction is assumed to be given beforehand. The viewing angle can be determined from the depth data. A Kalman filter is then used to track the liquid level and compensate for noise. Once it is detected that the liquid level has exceeded a user defined value, a stop signal is sent to terminate the pouring motion.

For detection and localizing of the user's mouth, we adopt a two step approach. In the first step, we segment the image based on the output of a face detection algorithm in order to extract the image region containing the user's mouth and eyes. Afterwards, we detect the position of the mouth of the user, considering only the obtained image patch. 
Regarding the mouth orientation, we additionally consider the position of the eyes in order to obtain a more robust estimation of the face orientation, hence compensating for slightly changing angles of the head. 
The face, mouth and eye detectors are implemented in OpenCV by applying an algorithm that uses Haar cascades~\cite{viola2001rapid, lienhart2002extended}.  

Table~\ref{tab:exp_drink} shows the averaged results for the experiment. Here, only 3.75\% of the 160 scheduled actions had to be repeated in order to complete the task successfully. In one run, plan recovery was not possible leading to abortion of the task. Thus, our system achieved in total a success rate of 90\% for the drinking task. Planning and execution required on average 545.56$\pm$67.38\,s. For the evaluation of the liquid level detection approach, we specified a desired fill level and executed 10 runs of the pour action. The resulting mean error and standard deviation is 6.9$\pm$8.9\,mm. In some instances the bottle obstructed the camera view, resulting in poor liquid level detection and a higher error. 

\section{Conclusions}
In this paper, we presented a thought-controlled mobile robotic service assistant, capable of successfully performing complex tasks, including close range interaction with the user, in a continuously changing environment to increase the independence of severely paralyzed patients. Through the use of a high-level planner as an intermediate layer between user and autonomous mobile robotic service assistant, we overcome the curse of dimensionality typically encountered in non-invasive BCI control schemes, thus opening up new perspectives for human-robot interaction scenarios. 

\bibliographystyle{IEEEtran}
\bibliography{literature}

\end{document}